%% file: main.tex
\documentclass[letterpaper, 10 pt, conference]{ieeeconf}
\IEEEoverridecommandlockouts
\usepackage{balance}
\usepackage{cite}
\usepackage{amsmath,amssymb,amsfonts,bbm,mathtools}
\usepackage{hyperref}

\usepackage[ruled, linesnumbered]{algorithm2e}
\SetAlCapNameFnt{\footnotesize}
\SetAlCapFnt{\footnotesize}

\usepackage{graphicx,subcaption}
\usepackage{textcomp}
\usepackage{xcolor}
\usepackage{comment}
\usepackage{multirow}
\def\BibTeX{{\rm B\kern-.05em{\sc i\kern-.025em b}\kern-.08em
    T\kern-.1667em\lower.7ex\hbox{E}\kern-.125emX}}

\usepackage{booktabs}

\newcommand{\methodShortName}{SURESTEP}

\DeclareMathOperator*{\argmin}{arg\,min}
    
\begin{document}

\title{\LARGE \bf \methodShortName{}: An Uncertainty-Aware Trajectory Optimization Framework to Enhance Visual Tool Tracking for \\Robust Surgical Automation}

\author{
Nikhil U. Shinde$^{*,1}$, Zih-Yun Chiu$^{*,1}$, Florian Richter$^{1}$, Jason Lim$^{1}$, Yuheng Zhi$^{1}$ \\Sylvia Herbert$^{2}$, and Michael C. Yip$^{1}$ \IEEEmembership{Senior Member, IEEE}
\thanks{\footnotesize $^*$ These authors contributed equally to this work.}
\thanks{
$^1$Nikhil U. Shinde, Zih-Yun Chiu, Florian Richter, Jason Lim, Yuheng Zhi, and Michael C. Yip are with the Electrical and Computer Engineering Dept., University of California San Diego, La Jolla, CA 92093 USA. {\tt\small \{nshinde, zchiu, frichter, jkl009, yzhi, yip\}@ucsd.edu}}%
\thanks{
$^2$Sylvia Herbert is with the Mechanical and Aerospace Engineering Dept., University of California San Diego, La Jolla, CA 92093 USA. {\tt\small sherbert@ucsd.edu}}%
}

\maketitle
\begin{abstract} 
\input{texfiles/Abstract}
\end{abstract}

\section{Introduction}
\input{texfiles/Introduction}

\section{Related Works}
\input{texfiles/RelatedWorks}

\section{Methodology}
\input{texfiles/Methods}

\section{Experiments and Results}
\input{texfiles/Experiments}

\section{Discussion and Conclusion}
\input{texfiles/conclusion}

\balance
\bibliographystyle{ieeeconf}
\bibliography{ref}

\end{document}

%% file: texfiles/Abstract.tex
Inaccurate tool localization is one of the main reasons for failures in automating surgical tasks. 
Imprecise robot kinematics and noisy observations caused by the poor visual acuity of an endoscopic camera make tool tracking challenging. 
Previous works in surgical automation adopt environment-specific setups or hard-coded strategies instead of explicitly considering motion and observation uncertainty of tool tracking in their policies. 
In this work, we present \methodShortName{}, an uncertainty-aware trajectory optimization framework for robust surgical automation. 
We model the uncertainty of tool tracking with the components motivated by the sources of noise in typical surgical scenes. 
Using a Gaussian assumption to propagate our uncertainty models through a given tool trajectory, \methodShortName{} provides a general framework that minimizes the upper bound on the entropy of the final estimated tool distribution. 
We compare \methodShortName{} with a baseline method on a real-world suture needle regrasping task under challenging environmental conditions, such as poor lighting and a moving endoscopic camera. 
The results over 60 regrasps on the da Vinci Research Kit (dVRK) demonstrate that our optimized trajectories significantly outperform the un-optimized baseline.

%% file: texfiles/Introduction.tex
Surgical automation can potentially revolutionize the consistency and accessibility of healthcare. 
Automating routine tasks during minimally invasive surgeries (MIS) can help reduce a surgeon's fatigue~\cite{hubens2003performance}. 
In addition, automation of standard procedures can help bring surgeries to underprivileged areas that lack medical expertise~\cite{khubchandani2018geographic}. 
Over the past decade, research on automating surgical subtasks has been done on robotic platforms such as the da Vinci Research Kit (dVRK)~\cite{kazanzides2014open}, including suturing~\cite{iyer2013single,sen2016automating,pedram2020autonomous,schwaner2021autonomous}, blood suction~\cite{richter2021autonomous, huang2021model}, and tissue dissection~\cite{oh2023framework}. 

The perception and localization of surgical tools is key to successful surgical automation. 
In MIS, this is primarily done through an endoscopic camera. 
Prior work such as~\cite{hao2018vision,lu2021super,richter2021robotic,dambrosia2024robust} focuses on accurately tracking the pose of surgical tools from endoscopic images. 
Automation work such as~\cite{d2018automated,chiu2021bimanual,wilcox2022learning} relies on good pose estimation as inputs to execute policies. 
Thus, automation often fails when poor views of tools prevent the detection of distinguishable features and hinder accurate pose estimation. 
These poor views are the result of edge distortions, dim lighting, and a lack of sharpness stemming from the endoscopic camera's poor visual acuity. 
In addition, while robot kinematics aid in tracking, surgical robots are particularly prone to the effects of cable stretch and hysteresis~\cite{miyasaka2020modeling,hwang2020efficiently} that severely degrade accuracy when using kinematics. 
These factors make pose estimation in surgical scenes particularly challenging. 

\begin{figure}[t]
    \centering
    \includegraphics[width=\linewidth]{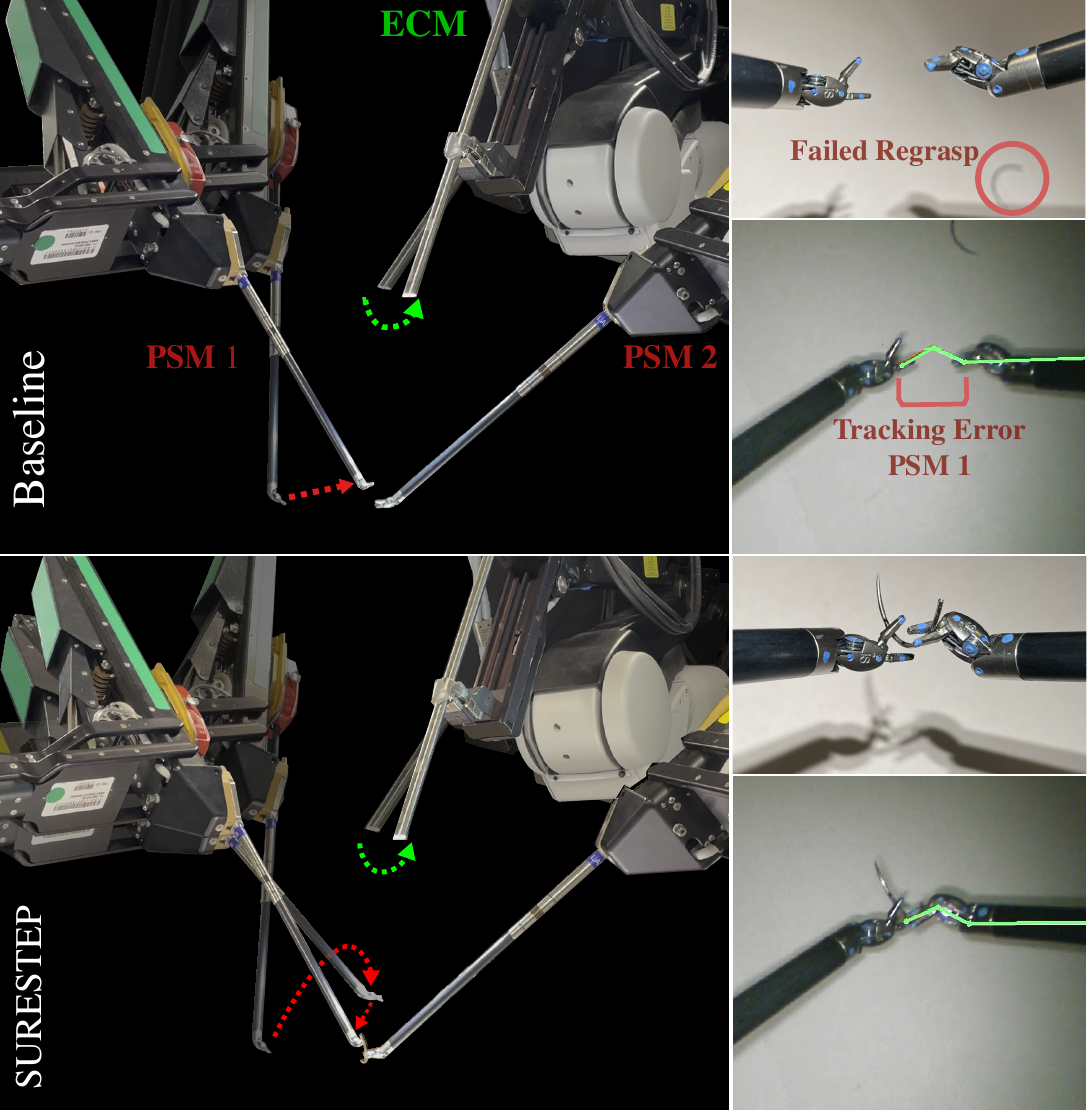}
    \caption{  
    Visualization of the baseline's (top) and \methodShortName{}'s (bottom) trajectories on the dVRK.  
    The first patient side manipulator (PSM 1) moves along a trajectory (red-dot arrows) to regrasp the needle held in PSM 2. 
    The PSMs and the needle are tracked using an endoscopic camera manipulator (ECM). 
    In our experiments, the ECM also moves along a given trajectory (green-dot arrows), adding extra noise to tool tracking. 
    The baseline trajectory fails to regrasp the needle due to significant tracking noise. 
    \methodShortName{} considers motion and observation uncertainty during trajectory optimization, improving tool tracking and achieving successful regrasps. 
    }
    \vspace{-5mm}
    \label{fig:cover_figure}
\end{figure}

To tackle visual challenges, past automation approaches introduce specific setups or policies for better tool visibility. 
For example, \cite{d2018automated} requires a calibrated 3D checkerboard workspace to pick up suture needles successfully. 
\cite{wilcox2022learning} moves the suture needle to a fixed home position and performs preset rotation policies until the needle is localized enough.
This policy is specific to free-space regrasping and only applies to a suture needle instead of other surgical tools.

Another line of research has looked into automating the endoscopic camera's movement for good visibility~\cite{eslamian2016towards,ji2018learning,moccia2023autonomous}. 
These works generate camera motions that make surgical tools appear at the center of the field of view (FOV).
While we incorporate a similar strategy in our framework, these approaches do not consider adjusting the tool's trajectory for better visibility.
Furthermore, we consider the scenario where there is no control of the camera's motion, such as shared autonomy.
Nevertheless, our framework can also be utilized to optimize the camera's motion for enhanced tool visibility.

In this work, we introduce \textbf{S}urgical \textbf{U}ncertainty-aware \textbf{R}obust \textbf{ES}timation \textbf{T}raj\textbf{E}ctory optimization \textbf{P}rotocol (\textbf{SURESTEP}), a framework for uncertainty-aware trajectory optimization that leads to robust surgical task automation through enhanced tool tracking. 
Our approach optimizes any trajectory outputted from an existing policy so that it is robust to the motion and observation uncertainty of surgical tools commonly encountered during surgeries. 
We propose different components that model the sources of motion and observation uncertainties. 
Moreover, we propose an objective function that minimizes the uncertainty of the estimated tool pose after propagating the motion and observation uncertainty in the belief space. 
We show that this objective function minimizes the upper bound on the entropy of the tracked surgical state distribution at the end of the trajectory. 

We apply SURESTEP to a real-world suture needle regrasping task~\cite{chiu2021bimanual,wilcox2022learning}, which requires accurate pose tracking of the surgical manipulators and needles. 
We showcase that SURESTEP largely improves the success rate of needle regrasping on the dVRK, even under adverse conditions, such as when one arm is initially out of the FOV, the initial needle is poorly visible, or when the endoscopic camera moves (which causes a large amount of motion noise in tool tracking). 
A visualization of non-optimized and our optimized trajectories are shown in Fig.~\ref{fig:cover_figure}. 
To the best of our knowledge, this is the first work demonstrating successful surgical task automation under a moving endoscopic camera. 

%% file: texfiles/RelatedWorks.tex
\subsection{Surgical Task Automation}

In recent years, researchers have looked into automating different surgical procedures, including needle regrasping~\cite{chiu2021bimanual,wilcox2022learning}, needle picking, insertion, or pulling~\cite{d2018automated,zhong2019dual,ozguner2021visually,fozilov2023towards}, suturing~\cite{iyer2013single,sen2016automating,pedram2020autonomous,schwaner2021autonomous}, blood suction~\cite{richter2021autonomous}, and vascular shunt insertion~\cite{dharmarajan2023automating}.
However, the methods to automate these procedures usually do not consider uncertainty in surgical tool localization.
Without taking uncertainty into account, deploying these methods in real-world environments requires the robot to always stay at the center of the FOV, run in highly controllable environments, or follow hand-crafted trajectories to make the tools clearly visible in images.
In~\cite{wilcox2022learning}, the authors design a \textit{suture needle acquisition} stage, in which the surgical manipulator moves to a fixed home pose to present the whole needle to the camera with a fixed policy.
\methodShortName{} results in a similar strategy for suture needle regrasping by explicitly considering the uncertainty of the needle's pose. 
However, our framework can be flexibly used with various surgical tools and considers diverse sources of uncertainty in surgical scenes.

\subsection{Planning and Control under Uncertainty}

Several works have considered the influence of uncertainty arising from the motion and observation models in state tracking. 
LQG-MP~\cite{LQG-MP} evaluates a set of trajectories by propagating Gaussian uncertainty through each trajectory using an extended Kalman filter (EKF) and outputs the one with the least probability of failing. 
\cite{van2012motion} formulates their planning-under-uncertainty problem as a Partially Observable Markov Decision Process (POMDP) and finds a locally optimal path by iteratively optimizing a linear control policy over the belief space. 
POMCPOW~\cite{sunberg2018online} solves the POMDP problem in a one-dimensional space with continuous actions and observations, where the robot aims to reach the goal under motion and observation noise. 
How uncertainty evolves throughout a trajectory is also a main focus of research in mobile robotics~\cite{huang2005multi,clemens2016evidential}.
Nonetheless, these techniques of planning under uncertainty have yet to address the challenges in surgical task automation. 
We provide a trajectory optimization framework to minimize the uncertainties in surgical tool tracking.
We jointly consider the sources of uncertainty in surgical scenes that were separately observed in former literature~\cite{ali2020supervised,eslamian2016towards,ji2018learning,eslamian2020development,moccia2023autonomous,ou2023robot,lin2023autonomous,reiter2012feature,ye2016real,hao2018vision,li2020super,richter2021robotic,lu2021super,dambrosia2024robust}, but not addressed together, to improve surgical task automation. 

%% file: texfiles/Methods.tex
\label{sec:methods}

We formulate the problem to optimize a given robot trajectory under uncertainty to minimize the expected distance between the final robot state and the desired goal.
This formulation can be written explicitly as:  
\begin{equation} 
\label{eq:problem_formulation_argmin}
\begin{aligned}
   \argmin_{\mathbf{u}_{1:T}} \quad & \mathbb{E}\left[\left\lVert \mathbf{x}_{T} - \mathbf{x}^{G} \right\rVert_{2}^{2}\right] \\
   \textrm{s.t.} \quad & \mathbf{x}_t = m(\mathbf{x}_{t-1}, \mathbf{u}_{t-1}, \mathbf{w}_{t-1}) \\
   &\mathbf{z}_t = h(\mathbf{x}_{t},  \mathbf{v}_t) \\
   &\mathbb{E}[\mathbf{x}_T] = \mathbf{x}^G
\end{aligned}
\end{equation}
$\mathbf{u}_{1:T}$ is the robot control throughout the trajectory, $\mathbf{z}_{1:T}$ are the observations of the robot, and $\mathbf{x}_{1:T}$ are the robot states. 
The robot states are random variables with $\mathbf{x}_T$ as the final state with distribution $\mathcal{P}(\mathbf{x}_T | \mathbf{\mathbf{u}}_{1:T}, \mathbf{z}_{1:T})$, and $\mathbf{x}^G$ is the desired goal state. 
$m(\cdot), h(\cdot)$ are the motion and observation models, and $\mathbf{w}_{1:T}, \mathbf{v}_{1:T}$ are stochastic noise in the motion and observation models respectively.
Our objective is expected L2 minimization of the robot state at $T$ from a given goal $\mathbf{x}^{G}$.
Since we consider the desired goal to be known, we ensure the final robot state is unbiased around the goal by imposing the constraint, $\mathbb{E}[\mathbf{x}_{T}] = \mathbf{x}^{G}$. 
In the coming subsections, we cover the optimization details, including how to estimate the distribution of the robot state at time $t$, $\mathcal{P}(\mathbf{x}_t | \mathbf{u}_{1:t}, \mathbf{z}_{1:t})$, through an EKF using motion and observation models designed for surgical tools.

\subsection{Modeling Surgical Robot State Estimation}

A robot state in this work refers to the pose of the surgical tool being tracked. 
The robot state, $\mathbf{x}_{t} = (\mathbf{x}^p_t, \mathbf{x}^o_t)$, is a Gaussian random variable, where $\mathbf{x}^p_t$ is the position, and $\mathbf{x}^o_t$ is the axis-angle orientation of the surgical tool. 
We assume that our motion and observation models, $m, h$, are locally linearizable and the motion and observation noise, $\mathbf{w}_{t}, \mathbf{v}_{t}$ can be modeled with zero mean multivariate Gaussian distributions: 
$\mathbf{w}_{t} \sim \mathcal{N}(\mathbf{0}, \mathbf{W}_{t}), \mathbf{v}_{t} \sim \mathcal{N}(\mathbf{0}, \mathbf{V}_{t})$.
Using these assumptions, we can model the evolution of
$\mathcal{P}(\mathbf{x}_t | \mathbf{u}_{1:t}, \mathbf{z}_{1:t})$
through the trajectory using an EKF. 
The predict step of the EKF models the evolution of our distribution after an action: 
\begin{equation}
\label{eq:predict_step}
    \begin{aligned}
        \boldsymbol{\mu}_{t+1|t} & = m(\boldsymbol{\mu}_{t|t}, \mathbf{u}_{t}, \mathbf{0})\\
        \boldsymbol{\Sigma}_{t+1|t} & = \mathbf{F}_{t} \boldsymbol{\Sigma}_{t|t} \mathbf{F}_{t}^{\top} + \mathbf{Q}_{t} \mathbf{W}_{t} \mathbf{Q}^{\top}_{t}
    \end{aligned}
\end{equation}
where $\mathbf{F}_{t} = \frac{\partial m}{\partial \mathbf{x}}(\boldsymbol{\mu}_{t|t}, \mathbf{u}_{t}, \mathbf{0})$, $\mathbf{Q}_{t} = \frac{\partial m}{\partial \mathbf{w}}(\boldsymbol{\mu}_{t|t}, \mathbf{u}_{t}, \mathbf{0})$, and $\mathbf{W}_{t}$ is the covariance of the motion noise at time $t$. 
The update step models how our distribution changes with an observation of the true state: 
\begin{equation} 
\label{eq:update_step}
    \begin{aligned}
        & \boldsymbol{\mu}_{t+1|t+1} = \boldsymbol{\mu}_{t+1|t} + \mathbf{K}_{t+1|t}(\mathbf{z}_{t+1} - h(\boldsymbol{\mu}_{t+1|t}, \mathbf{0})) \\
        & \boldsymbol{\Sigma}_{t+1|t+1} = (\mathbf{I} - \mathbf{K}_{t+1|t}\mathbf{H}_{t+1|t}) \boldsymbol{\Sigma}_{t|t} \\
        & \mathbf{K}_{t+1|t} = \boldsymbol{\Sigma}_{t+1|t}\mathbf{H}^{\top}_{t+1}(\mathbf{H}_{t+1}\boldsymbol{\Sigma}_{t+1|t}\mathbf{H}_{t+1}^{\top} + \mathbf{R}_{t+1}\mathbf{V}_{t+1}\mathbf{R}_{t+1}^{\top})
    \end{aligned}
\end{equation}
where $\mathbf{H}_{t+1} = \frac{\partial h}{\partial \mathbf{x}}(\boldsymbol{\mu}_{t+1|t}, \mathbf{0})$, $\mathbf{R}_{t+1} = \frac{\partial h}{\partial \mathbf{v}}(\boldsymbol{\mu}_{t+1|t}, \mathbf{0})$, 
$\mathbf{z}_{t+1} = h(\mathbf{x}_{t+1}, \mathbf{v}_{t+1})$ is an observation with noise $\mathbf{v}_{t+1} \sim \mathcal{N}(\mathbf{0}, \mathbf{V}_{t+1})$, and $\mathbf{K}_{t+1|t}$ is referred to as the Kalman Gain. 

The EKF provides a deterministic model of the belief dynamics, i.e., how our mean and covariance will evolve over a trajectory. 
We can use this model to evaluate and optimize a trajectory. 
When the predict and update steps in Equations (\ref{eq:predict_step}) and (\ref{eq:update_step}) propagate the distribution forward, we assume that $\mathbf{x}_{t+1} = m(\boldsymbol{\mu}_{t|t}, \mathbf{u}_{t}, \mathbf{0})$. 
We also assume maximum-likelihood observations~\cite{platt2010belief}, which are sampled at the current mean estimate of the robot state, i.e., $\mathbf{z}_{t+1} = h(\boldsymbol{\mu}_{t+1|t}, \mathbf{0})$.
This allows us to directly compute the expected distribution of the final robot state, $\mathbf{x}_{T} \sim \mathcal{N}(\boldsymbol{\mu}_{T|T}, \boldsymbol{\Sigma}_{T|T})$.
We use this deterministic computation to calculate the loss in Equation (\ref{eq:problem_formulation_argmin}) and directly optimize our trajectory in the belief space.

\subsection{Motion and Observation Models for Surgical Tools}

\begin{figure}[t!]
    \vspace{1.5mm}
    \centering
    \includegraphics[width=\linewidth]{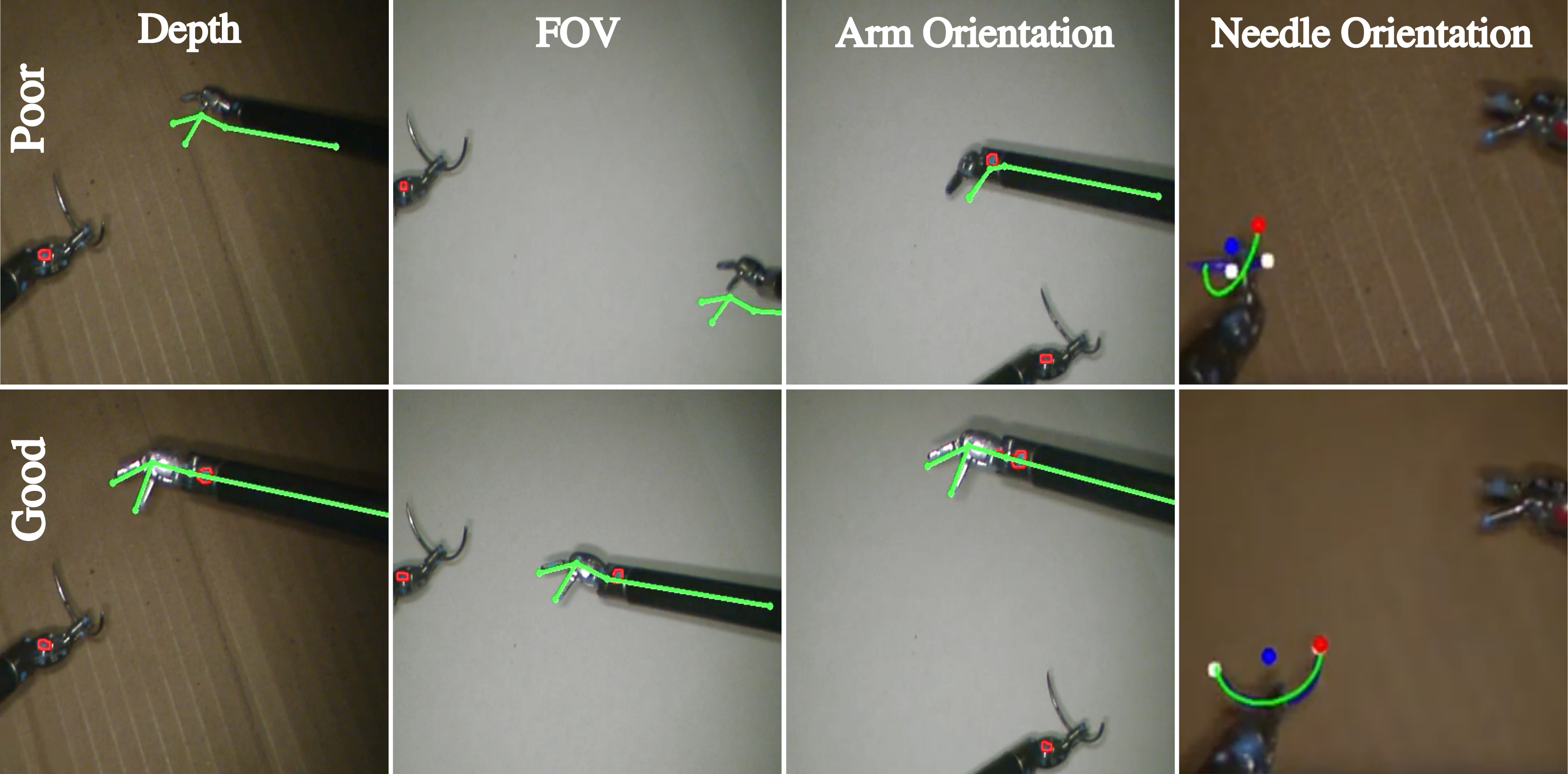 }
    \caption{ 
    Tracking results on the surgical manipulators and suture needles from an endoscopic camera using~\cite{richter2021robotic,chiu2022markerless,chiu2023real}. 
    The green curves show the tracked tool poses. 
    Each column highlights a source of observational uncertainty that impacts tool tracking. 
    }
    \vspace{-2mm}
    \label{fig:method-different-obs}
\end{figure}

Tool tracking for surgical subtasks relies on both robot kinematics as well as visual information from the surgical endoscope. 
When designing the motion and observation models to propagate the belief in the robot state, we should account for the challenges of tracking surgical tools. 

For motion models, we can directly use the robot kinematics to model the change in the surgical tool state after an action. 
However, joint encoder readings at the motors do not translate precisely to changes in robot pose due to deteriorated kinematics from cable stretch and hysteresis on cable-driven surgical manipulators~\cite{miyasaka2020modeling,hwang2020efficiently}.
We capture these effects through the motion noise models. 

We separate the motion model noise into position- and orientation-based components, $\mathbf{w}_{t} = \mathbf{w}^{p}_{t} + \mathbf{w}^{o}_{t}$, and propose to model their uncertainty to be proportional to the distance between waypoints, as larger motions result in greater uncertainty.
The position-based motion noise is sampled from $\mathcal{N}(\mathbf{0}, \mathbf{W}^{p}_{t}(\mathbf{u}_{t}))$:  
\begin{equation}
    \label{eq:positional_motion_noise}
    \mathbf{W}^{p}_{t}(\mathbf{u}_{t}) = \left\lVert \mathbf{x}^p_{t+1} - \mathbf{x}^p_{t} \right\rVert_{2}^{2} \mathbf{W}^{p,0}_{t}
\end{equation}
where $\mathbf{W}^{p,0}_{t}$ is a base covariance matrix.
The orientation-based motion noise is sampled from $\mathcal{N}(\mathbf{0}, \mathbf{W}^{o}_{t}(\mathbf{u}_{t}))$: 
\begin{equation}
    \label{eq:orientational_motion_noise}
    \mathbf{W}^{o}_{t}(\mathbf{u}_{t}) = \left(A(\mathbf{x}^o_{t}, \mathbf{x}^o_{t+1})\right)^2 \mathbf{W}^{o,0}_{t} 
\end{equation}
$A(\mathbf{x}^o_{t}, \mathbf{x}^o_{t+1})$ computes the angle difference between the trajectory waypoints at time $t$ and $t+1$. 
The full pose motion noise thus becomes: $\mathbf{w}_{t} = \mathbf{w}^{p}_{t}(\mathbf{u}_{t}) + \mathbf{w}^{o}_{t}(\mathbf{u}_{t}) \sim \mathcal{N}\left(\mathbf{0}, \mathbf{W}^{p}_{t}(\mathbf{u}_{t}) + \mathbf{W}^{o}_{t}(\mathbf{u}_{t})\right)$. 

The tool state predicted by the motion model is updated using observations from the surgical endoscopic camera.  
Prior methods extract surgical-tool segmentation or keypoints from the images as observations to update the current mean estimate~\cite{reiter2012feature,ye2016real,hao2018vision,li2020super,lu2021super,richter2021robotic,dambrosia2024robust}. 
However, these vision-based detectors can yield poor observations when views of the surgical tool are far from ideal detection circumstances. 
We capture these effects through our observation noise model: $\mathbf{v}_{t}(\mathbf{x}_{t}) \sim \mathcal{N}(\mathbf{0}, \mathbf{V}_{t}(\mathbf{x}_{t}))$, which is a function of the state. 

Fig.~\ref{fig:method-different-obs} shows how different factors affect surgical tool detection in endoscopic images.  
These factors are also separately discussed in prior work~\cite{ali2020supervised,eslamian2016towards,ji2018learning,eslamian2020development,moccia2023autonomous,ou2023robot,lin2023autonomous,reiter2012feature,ye2016real,hao2018vision,li2020super,richter2021robotic,lu2021super,dambrosia2024robust}. 
Thus, we propose $3$ components for the observation noise: depth-based $\mathbf{v}^{d}_{t}$, field-of-view (FOV) based $\mathbf{v}^{f}_{t}$, and orientation-based observation noise $\mathbf{v}^{o}_{t}$.
The final observation noise is the sum of the individual noise components, i.e., $\mathbf{v}_{t} = \mathbf{v}^{d}_{t} + \mathbf{v}^{f}_{t} + \mathbf{v}^{o}_{t}$, which are all defined as zero-mean, multivariate Gaussians. 
Although here we introduce the motion and observational uncertainties commonly seen in surgical scenes, \methodShortName{} is a general framework within which the motion and observation models can be extended to fit diverse surgical tasks, tools, and detection models.

\subsubsection{Depth-based observation noise}
As shown in the first column of Fig.~\ref{fig:method-different-obs}, when the tool is too far from the camera, poor lighting in combination with the poor visual acuity of the endoscopic camera can make it hard to detect distinguishable features for good pose estimation. 
Meanwhile, when the tool is too close to the camera, it is subject to excessive reflections from the light source attached to the camera.
Thus, the depth-based observation noise is modeled as $\mathbf{v}^{d}_{t}(\mathbf{x}_{t}) \sim \mathcal{N}(\mathbf{0}, \mathbf{V}^{d}_{t}(\mathbf{x}_{t}))$, and the covariance is proportional to the distance from an ideal detection depth in the camera frame: 
\begin{equation}
    \mathbf{V}^{d}_{t}(\mathbf{x}_{t}) = \left( d^{c}_{t}(\mathbf{x}_{t}) - d^*_{t} \right)^{2} \mathbf{V}^{d,0}_{t}
\end{equation} 
Here, $d^{c}_{t}(\mathbf{x}_{t})$ is the depth of the robot state in the camera frame at waypoint $\mathbf{x}_{t}$, $d^*_{t}$ is the ideal depth in the camera frame, and $\mathbf{V}^{d,0}_{t}$ is the base depth covariance matrix. 

\subsubsection{FOV-based observation noise}
The second column in Fig.~\ref{fig:method-different-obs} shows that the surgical tool's location in the camera's FOV also has a large impact on the detections. 
An endoscopic camera's FOV is narrow, with the edges of the FOV suffering from large distortions. 
This makes the detections of the surgical tool far from the center of the FOV subject to large noise. 
Thus, we model the FOV-based observation noise as $\mathbf{v}^{f}_{t}(\mathbf{x}_{t}) \sim \mathcal{N}(\mathbf{0}, \mathbf{V}^{f}_{t}(\mathbf{x}_{t}))$, and the covariance is proportional to the distance of the surgical tool from the image's center: 
\begin{equation}
    \mathbf{V}^{f}_{t}(\mathbf{x}_{t}) = \left\lVert \mathbb{I}(\mathbf{x}_{t}) - \mathbb{I}_c \right\rVert^{2}_2 \mathbf{V}^{f,0}_{t}
\end{equation}
$\mathbb{I}(\mathbf{x}_{t})$ is the projection of $\mathbf{x}_{t}$ onto the image plane, $\mathbb{I}_c$ is the image's center, and $\mathbf{V}^{f,0}_{t}$ is the base FOV covariance matrix. 

\subsubsection{Orientation-based observation noise}
From the third and fourth columns of Fig.~\ref{fig:method-different-obs}, we can see that the detection quality is also dependent on the orientation of the tool. 
In some orientations, the tool or its keypoints can be occluded, worsening the tracking results. 
Thus, we model the orientation-based observation noise as $\mathbf{v}^{o}_{t}(\mathbf{x}_{t}) \sim \mathcal{N}(\mathbf{0}, \mathbf{V}^{o}_{t}(\mathbf{x}_{t}))$, with the covariance proportional to the difference in angle from a desirable orientation where the features of the tool are mostly in view: 
\begin{equation}
    \mathbf{V}^{o}_{t}(\mathbf{x}_{t}) = \left(1 - \frac{\mathbf{x}^{o}_t \cdot \mathbf{o}^{*}}{\lVert \mathbf{x}^{o}_t \rVert \lVert \mathbf{o}^{*} \rVert}\right)^{2} \mathbf{V}^{o,0}_{t}
\end{equation}
Here, $\mathbf{o}^{*}$ is the optimal orientation for detections, and $\mathbf{V}^{o,0}_{t}$ is the base orientation covariance matrix.

\begin{figure}
    \centering
    \includegraphics[width=\linewidth]{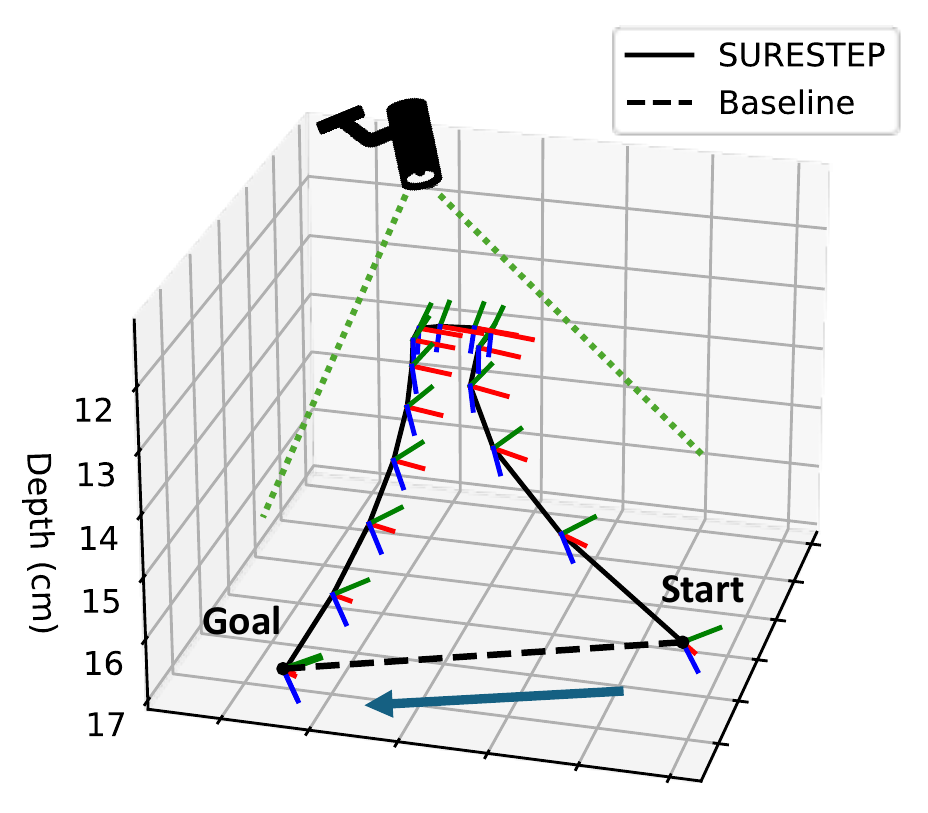}
    \caption{
    Trajectories before and after optimization through \methodShortName{}. 
    The optimized trajectory demonstrates the expected behavior of a tool moving closer to the camera and the center of FOV to reduce tracking uncertainty before returning to the desired goal pose.}
    \label{fig:opt_traj_ours}
\end{figure}

\subsection{Optimization}

With the motion and observation models and the deterministic belief dynamics of the EKF in Equations (\ref{eq:predict_step}) and (\ref{eq:update_step}), we have an explicit expression of the final robot state distribution, $\mathbf{x}_{T} \sim \mathcal{N}(\boldsymbol{\mu}_{T|T}, \boldsymbol{\Sigma}_{T|T})$. 
Thus, the expectation of the objective function in Equation (\ref{eq:problem_formulation_argmin}) has a closed-form expression that we can optimize directly.
We derive this expression by rewriting Equation (\ref{eq:problem_formulation_argmin}) as 
\begin{equation}
\label{eq:argmin_rewrite_1}
    \begin{aligned}
        & \mathbb{E}\left[\left\|\mathbf{x}_{T} - \mathbf{x}^{G}\right\|_{2}^{2}\right] \\ 
        =\ & \mathbb{E}\left[ \mathbf{x}_{T}^{\top}\mathbf{x}_{T} \right] - 2\mathbb{E}\left[ \mathbf{x}_{T}^{\top} \right] \mathbf{x}^{G} + \mathbf{x}^{G\top}\mathbf{x}^{G} \\
        =\ & \mathbb{E}\left[ \mathbf{x}_{T}^{\top}\mathbf{x}_{T} \right] - \mathbf{x}^{G\top}\mathbf{x}^{G}
    \end{aligned}
\end{equation}
The last equality comes from our constraint, $\mathbb{E}\left[ \mathbf{x}_{T} \right] = \mathbf{x}^{G}$, in Equation (\ref{eq:problem_formulation_argmin}). 
We can then use the following relationship between $\text{Tr}(\boldsymbol{\Sigma}_{T|T})$ and $\mathbb{E}\left[\mathbf{x}_{T}^{\top}\mathbf{x}_{T}\right]$, where $\text{Tr}$ is the trace: 
\begin{equation}
\label{eq:cov_trace_relation}
    \begin{aligned}
        \text{Tr}(\boldsymbol{\Sigma}_{T|T}) & = \text{Tr}\left(\mathbb{E}\left[ (\mathbf{x}_{T}-\boldsymbol{\mu}_{T|T})(\mathbf{x}_{T}-\boldsymbol{\mu}_{T|T})^{\top} \right]\right) \\
        & = \mathbb{E}\left[\mathbf{x}_{T}^{\top}\mathbf{x}_{T}\right] - \boldsymbol{\mu}_{T|T}^{\top}\boldsymbol{\mu}_{T|T} \\
    \end{aligned}
\end{equation}
Substituting $\mathbb{E}\left[\mathbf{x}_{T}^{\top}\mathbf{x}_{T}\right] = \text{Tr}(\boldsymbol{\Sigma}_{T|T}) + \boldsymbol{\mu}_{T|T}^{\top}\boldsymbol{\mu}_{T|T}$ into Equation (\ref{eq:argmin_rewrite_1}), the optimization problem in Equation (\ref{eq:problem_formulation_argmin}) becomes 
\begin{equation}
    \label{eq:real_objective_function}
    \begin{aligned}
        & \argmin_{\mathbf{u}_{1:T}} \text{Tr}(\boldsymbol{\Sigma}_{T|T}) + \boldsymbol{\mu}_{T|T}^{\top}\boldsymbol{\mu}_{T|T} - \mathbf{x}^{G\top}\mathbf{x}^{G} \\
        = & \argmin_{\mathbf{u}_{1:T}} \text{Tr}(\boldsymbol{\Sigma}_{T|T}) \\
    \end{aligned}
\end{equation}
Thus, we minimize the trace of the final covariance matrix. 

\begin{table*}[t!]
\vspace{1.5mm}
\centering
\caption{
Ablation study results over 100 initial configurations and 50 noisy rollouts for each configuration. 
The numbers are relative to the baseline, with lower values indicating better performance. 
}
\label{tab:ablation_study_with_noise}
\begin{tabular}{c cc cc cc cc}
    \toprule
     \multirow{3}{*}{Metric} & \multicolumn{2}{c}{Position diff.} & \multicolumn{2}{c}{Orientation diff.} & \multicolumn{2}{c}{Trace of $\boldsymbol{\Sigma}_{T|T}$} & \multicolumn{2}{c}{Entropy of $\mathbf{x}_T$} \\
     \cmidrule(lr){2-3}\cmidrule(lr){4-5}\cmidrule(lr){6-7}\cmidrule(lr){8-9}
     & mean & std & mean & std & noisy & max likelihood & noisy & max likelihood \\
    \midrule
    Baseline & 1 & 1 & 1 & 1 & 1 & 1 & 1 & 1 \\
    SURESTEP (all) & \textbf{0.4272} & \textbf{0.5425} & \textbf{0.1747} & \textbf{0.4633} & \textbf{0.1597} & \textbf{0.1819} & \textbf{0.3677} & \textbf{0.5078} \\
    SURESTEP (no pose loss) & \textbf{0.412} & \textbf{0.5205} & \textbf{0.16} & \textbf{0.3622} & \textbf{0.1494} & \textbf{0.1401} & \textbf{0.3265} & \textbf{0.4311} \\
    No depth noise & 0.5019 & 0.553 & 0.2419 & 0.4994 & 0.2188 & 0.3183 & 0.5046 & 0.6658 \\
    No FOV noise & 0.4743 & 0.5771 & 0.2136 & 0.5098 & 0.2033 & 0.246 & 0.4636 & 0.6016 \\
    No orientation noise & 0.9862 & 0.978 & 0.8929 & 0.9425 & 0.9375 & 0.9441 & 0.9723 & 0.9773 \\
    \bottomrule
    \end{tabular}
    \vspace{-2mm}
\end{table*}

We can show that this minimization also seeks to decrease an upper bound on the entropy of the final distribution of our robot state $\mathbf{x}_{T}$ in the case that the final covariance $\boldsymbol{\Sigma}_{T|T}$ is positive definite. 
The entropy of a multivariate Gaussian distribution is: 
\begin{equation}
\label{eq:entropy}
    H(\mathbf{x}) = c + \frac{1}{2}\ln(|\boldsymbol{\Sigma}|) = c + \frac{1}{2}\sum_{i}\ln(\lambda_{i})
\end{equation}
where $c$ is some positive constant, and $\lambda_i$ are the eigenvalues of $\boldsymbol{\Sigma}$. 
Thus, the entropy of the final robot state $H(\mathbf{x}_{T}) \propto \ln(|\boldsymbol{\Sigma}_{T|T}|) + c$. 
Here, we consider positive definite $\boldsymbol{\Sigma}_{T|T}$ since zero eigenvalues make $H(\mathbf{x}_{T})$ ill defined. 
Then, we can show that minimizing the trace of $\boldsymbol{\Sigma}_{T|T}$ implies minimizing an upper bound of the entropy of $\mathbf{x}_T$. 

We start by showing the relationship between the trace and log determinant of a positive definite matrix. 
Let $\lambda_i = e^{q_i}$, 
\begin{equation}
    \ln(|\boldsymbol{\Sigma}|) = \ln\left(\prod_{i}\lambda_i\right) = \sum_{i} \ln(\lambda_i) = \sum_{i} q_{i} 
\end{equation}
We can substitute $\lambda_{i} = e^{q_i}$ since $\lambda_{i} > 0$ if the matrix is positive definite. 
From Bernoulli's inequality, we have $e^{x} \geq 1 + x$, and thus $e^{x} > x$. 
With this, we get: 
\begin{equation}
    \begin{split}
        \ln(|\boldsymbol{\Sigma}|) &= \sum_{i} q_{i} < \sum_{i} e^{q_{i}} \\
        \ln(|\boldsymbol{\Sigma}|) & < \sum_{i} \lambda_{i} = \text{Tr}(\boldsymbol{\Sigma}) \\
    \end{split}
\end{equation}
Thus, $H(\mathbf{x}) < c + \frac{1}{2} \text{Tr}(\boldsymbol{\Sigma})$. 

We solve our optimization problem in Equation (\ref{eq:real_objective_function}) using a gradient-based approach. 
Given an initial trajectory, we pass it through the deterministic belief model of EKF and compute the loss, $\text{Tr}(\boldsymbol{\Sigma}_{T|T})$. 
The gradient-based approach iteratively modifies the difference between the trajectory's waypoints to decrease the loss until convergence. 
We enforce the constraint in our optimization, $\mathbb{E}[\mathbf{x}_{T}] = \mathbf{x}^{G}$, by fixing the final waypoint in our optimized trajectory to be $\mathbf{x}^{G}$. 
We refer to this final action as the \textit{clipping action}.
In addition, we impose a pose loss to help the optimization converge to a trajectory where the last action is not too large. 
Thus, the final objective function becomes: 
\begin{equation}
    \label{eq:final_objective_function}
    \text{Tr}(\boldsymbol{\Sigma}_{T|T}) + \left\lVert \mathbf{x}_T^p - \mathbf{x}_{T-1}^p \right\rVert_2 + A(\mathbf{x}_{T-1}^o, \mathbf{x}_T^o).
\end{equation}
Fig. \ref{fig:opt_traj_ours} shows a trajectory optimized by \methodShortName{}. 

%% file: texfiles/Experiments.tex
We implement our motion models, observation models, and EKF using PyTorch~\cite{paszke2019pytorch}. 
All orientation-related operations are implemented using PyPose~\cite{wang2023pypose}. 
This is done to ease gradient computation with auto differentiation. 
For optimization, we use limited memory BFGS (LBFGS)~\cite{liu1989limited} to minimize our loss function and let the optimization run for $50$ iterations. 
As is common practice in safety literature \cite{HJReachability, GP_reachability}, where worst-case bounds on noise and disturbance are used to provide safety guarantees, we optimize using worst-case assumptions on noise to ensure improved performance.
The constants used in optimization are as follows: $\mathbf{W}^{p,0}= 10^{-3}\mathbf{I}$, $\mathbf{W}^{o,0}= 10^{-3}\mathbf{I}$, $\mathbf{V}^{d,0}= 10^{-1}\mathbf{I} $, $\mathbf{V}^{f,0} = 10^{-2}\mathbf{I}$, and $\mathbf{V}^{o,0}= 5 \times 10^{-3}\mathbf{I}$, where $\mathbf{I} \in \mathbb{R}^{6 \times 6}$ is the identity matrix. 
We use a prior covariance of $10^{-2}\mathbf{I}$, and empirically set the optimal depth $d^{*}_{t} = 0.15$. 
All constants are in meters. 

In experiments, \methodShortName{} optimizes a trajectory generated by a baseline approach. 
This baseline does not consider uncertainty in state estimation and can be methods such as sampling-based motion planners or~\cite{chiu2021bimanual}. 
Here, we use path interpolation as our baseline since there are no obstacles in our environments, and interpolation provides the minimum path length between the start and goal poses.

\subsection{Simulation Experiments}

We perform an ablation study in simulation to demonstrate the effects of our proposed uncertainty components and pose loss from Section~\ref{sec:methods}. 
We compare the baseline, \methodShortName{}, and variants of \methodShortName{} that optimize trajectories by not considering the effects of some components: 
\begin{itemize}
    \item \textit{Baseline}: No optimization. 
    \item \textit{SURESTEP (all)}: Include all components in Section~\ref{sec:methods}. 
    \item \textit{SURESTEP (no pose loss)}: Optimize Equation (\ref{eq:real_objective_function}) instead of (\ref{eq:final_objective_function}).
    \item \textit{No depth noise}: $\mathbf{V}_t^d(\mathbf{u}_t) = \mathbf{0}$ in optimization. 
    \item \textit{No FOV noise}: $\mathbf{V}_t^f(\mathbf{u}_t) = \mathbf{0}$ in optimization. 
    \item \textit{No orientation noise}: $\mathbf{V}_t^o(\mathbf{u}_t) = \mathbf{0}$ in optimization. 
\end{itemize}
When evaluating the trajectories, we consider all sources of observation and motion noise, $\mathbf{v}_{t} \sim \mathcal{N}(\mathbf{0}, \mathbf{V}_{t}), \mathbf{w}_{t} \sim \mathcal{N}(\mathbf{0}, \mathbf{W}_{t})$.
We evaluate the baseline and optimized trajectories on 100 randomly initialized configurations; each of them is different in the needle or camera pose. 
These initial configurations are validated using CoppeliaSim~\footnote{https://coppeliarobotics.com/}. 
 
Fig.~\ref{fig:opt_traj_ours} visualizes trajectories before and after optimization through \methodShortName{}. 
Table~\ref{tab:ablation_study_with_noise} reports the mean and standard deviation of the positional and orientational distance between the actual and desired final pose. 
We also report the trace and entropy of the final covariance, $\boldsymbol{\Sigma}_{T|T}$, tracked by the EKF after a noisy rollout and under the maximum-likelihood assumption~\cite{platt2010belief}, i.e., no noise samples.
From the tracked trace and entropy with and without noise, we can see that the trend observed in the maximum-likelihood case, considered in optimization, aligns with the trend when noise actively affects a trajectory.
Note that we compare the performance of all methods \textit{relative to the baseline} in Table~\ref{tab:ablation_study_with_noise}.  
These relative values are calculated by scaling their original values: $\frac{y}{b}$ if $y$ and $b$ are positive, and $1 - \frac{y - b}{b}$ if $y$ and $b$ are negative (entropy), where $y$ is the un-scaled value of a metric, and $b$ is the un-scaled value of a baseline's metric.

The results show that our method, when considering all sources of observation noise, leads to a lower mean and standard deviation in the positional and orientational distance from the desired pose. 
In addition, our final tracked covariance achieves a smaller trace and entropy. 
This demonstrates that by considering uncertainty, we can find trajectories that increase our tracking confidence and the precision with which our trajectory leads to the desired goal. 

Omitting different observational uncertainties in optimization affects the performance of all metrics.  
While the results with no depth noise and no FOV noise are still apparently better than the baseline, the results with no orientation noise are only slightly better. 
Since the solution space of trajectories is large, without considering one source of uncertainty, the optimization can generate a trajectory poorly affected by this unconsidered factor. 
This indicates the importance of considering varying uncertainties during optimization. 

From Table~\ref{tab:ablation_study_with_noise}, the best results come from when we exclude the pose loss from the objective function, i.e., optimizing Equation (\ref{eq:real_objective_function}). 
Without considering the pose loss, the optimization can focus on reducing the trace while allowing the clipping action to enforce the final pose constraint. 
Given enough optimization iterations, the trajectory can converge smoothly to the goal. 
However, adding the pose loss in optimization can give additional benefits: 
It stabilizes the output trajectory and ensures that the clipping action stays reasonably small throughout optimization. 
Thus, if we are constrained by time and end the optimization early, the output trajectory will be good enough for a successful rollout.

\subsection{Real World Experiments}

\begin{table}[t!]
\vspace{1.5mm}
\centering
\caption{Success rate for needle regrasping on dVRK}
\label{tab:real-robot-success-rate}
\begin{tabular}{ccccccc}
    \toprule
    Type & \multicolumn{3}{c}{1} & 2 & 3 & 4 \\
    \cmidrule(lr){2-4}
    Env. & \multirow{2}{*}{Easy} & \multirow{2}{*}{Med.} & \multirow{2}{*}{Hard} & Dim & ECM & ECM \\
    Condition &  &  &  & light & moves (1) & moves (4) \\
    \midrule
    Baseline &          5/10 &          0/10 &          0/10 &          4/10 &          2/10 &           0/10 \\
    \methodShortName{} & \textbf{9/10} & \textbf{9/10} & \textbf{7/10} & \textbf{9/10} & \textbf{8/10} & \textbf{7/10} \\
    \bottomrule
    \end{tabular}
    \vspace{-3mm}
\end{table}

\begin{figure}[t!]
    \vspace{1.5mm}
    \centering
    \includegraphics[width=\linewidth]{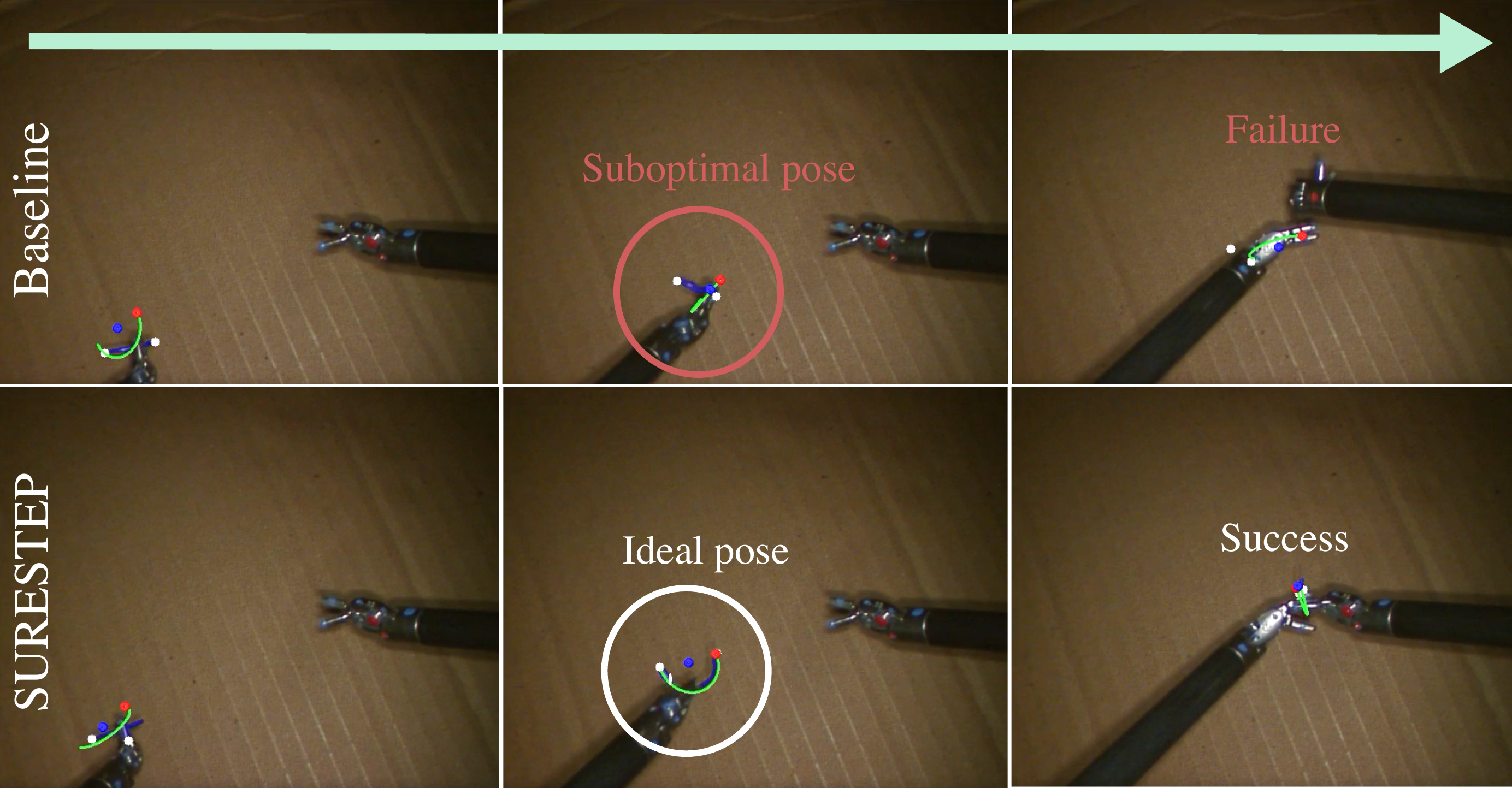}
    \caption{
    First-person view of a Type 2 trajectory, in which the needle arm moves to the regrasping arm. 
    Here, real-time needle tracking is required to perform a regrasp. 
    Note that dim lighting makes the visibility conditions challenging, so needle segmentation often fails. 
    The baseline trajectory cannot recover from inaccurate needle pose estimation, leading to a failed regrasp. 
    \methodShortName{}'s trajectory considers observational uncertainty and re-orients the needle for better pose estimation, hence succeeding in regrasping. 
    }
    \label{fig:ECM-fixed-needle-moves-first-person}
    \vspace{-1mm}
\end{figure}

We evaluate the non-optimized and optimized needle-regrasping trajectories in the real world on a dVRK~\cite{kazanzides2014open}. 
A suture needle with a 7mm or 11.5mm radius is initially grasped in a Large Needle Driver (LND) attached to one of the Patient Side Manipulator (PSM) arms from the dVRK. 
We refer to the PSM arm with a needle as the \textit{needle arm} and another PSM arm that performs the regrasping as the \textit{regrasping arm}. 
The PSM arms and suture needles are tracked from the dVRK's stereo endoscope using previous methods~\cite{richter2021robotic,chiu2022markerless,chiu2023real}, which give us the poses of end-effectors and needles in the camera frame. 
The stereo endoscope, which is 1080p and runs at 30 fps, is held by an endoscopic camera manipulator (ECM). 
The markerless needle detections are obtained in real-time using Cutie~\cite{cheng2023putting}, a video-object segmentation network, with an initial segmentation extracted by the Segment Anything Model (SAM)~\cite{kirillov2023segment}. 
We did not perform further tuning on these models to fit our environments. 
The surgical manipulator and needle tracking algorithms run at $20$ fps. 

We generated four different types of trajectories: 
\begin{enumerate}
    \item ECM is fixed. The regrasping arm moves to regrasp while the needle arm is fixed.
    \item ECM is fixed. The needle arm moves to regrasp while the regrasping arm is fixed. 
    \item ECM moves once at the beginning. Then, the regrasping arm moves to regrasp while the needle arm is fixed. 
    \item ECM moves throughout the trajectory. The regrasping arm moves to regrasp while the needle arm is fixed.
\end{enumerate}
The first two types of trajectories demonstrate that \methodShortName{} can optimize the trajectories of different surgical tools. 
We emphasize that the third and fourth types of trajectories are commonly seen in manual laparoscopic surgeries yet are challenging for a robot to perform successfully. 
In laparoscopic surgeries, camera assistants frequently follow a surgeon's commands to adjust the camera for a better field of view~\cite{omote1999self}. 
However, while moving the camera, a significant amount of noise will be present in videos due to motion blur or shaking of the camera manipulator, causing the uncertainty in state estimation to largely increase. 

For all types of trajectories, we first captured one to three initial configurations of the regrasping arm, the needle arm, and the needle using the tool- and needle-tracking methods. 
Given these initial configurations, we generated the baseline and our optimized trajectories. 
Then, for each trajectory, we ran ten trials to evaluate the robustness of each method.
During the fourth type of trajectory, we re-optimized the trajectory after each ECM movement. 
Note that most of the trajectories we ran are subject to challenging environmental conditions, e.g., one arm is out of the FOV, the scene is with dim light, or the ECM moves.

\begin{figure}[t!]
    \vspace{1.5mm}
    \centering
    \includegraphics[width=\linewidth]{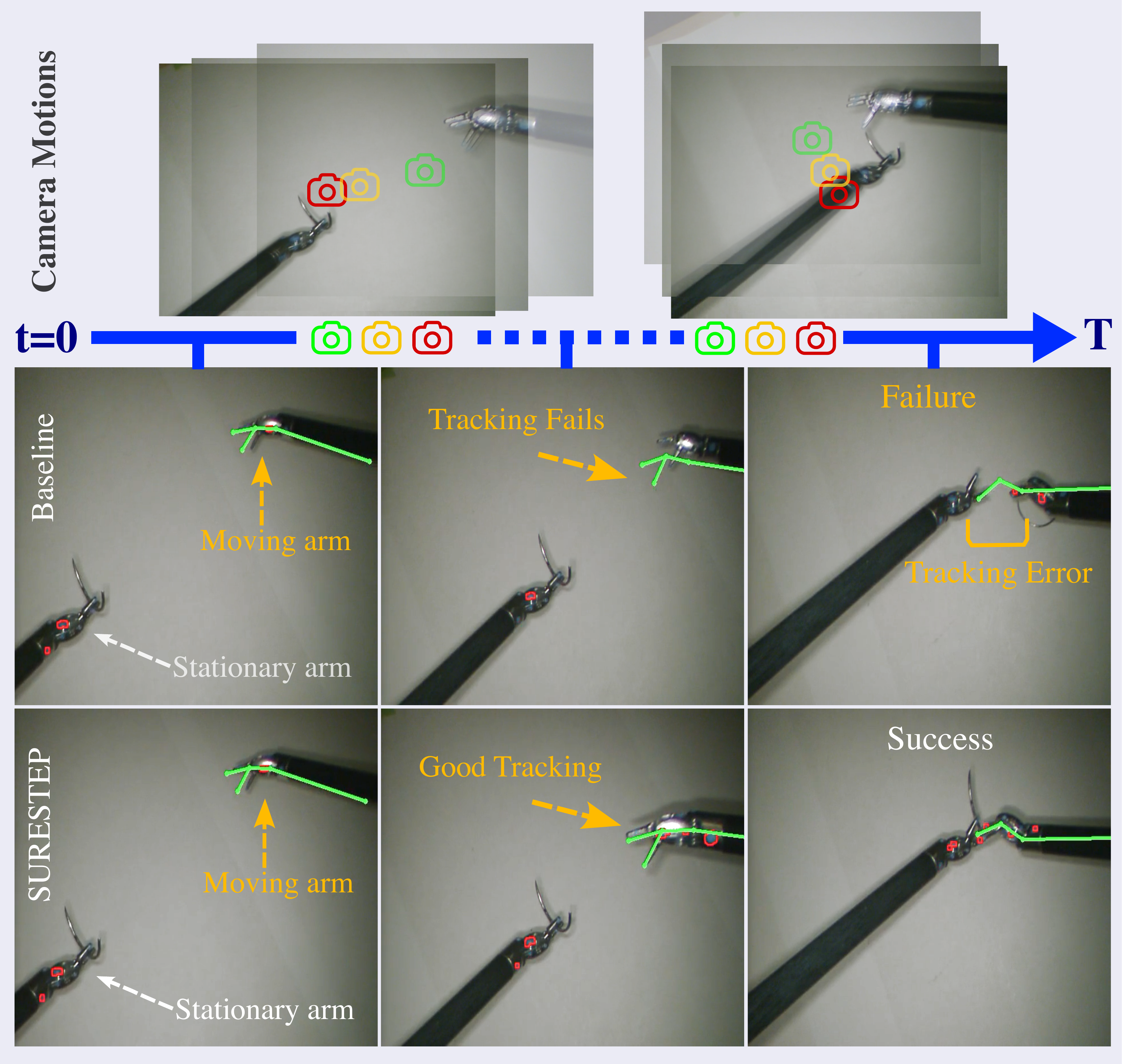}
    \caption{
    First-person view of a Type 4 trajectory, in which the ECM moves in the middle of the regrasping arm's trajectory.  
    The top row depicts each camera movement, from the frame with the green icon to the one with the red icon. 
    The baseline trajectory fails to regrasp since compounding noise from the ECM's motion and other sources leads to inaccurate tool tracking. 
    \methodShortName{}'s trajectory considers this uncertainty and is re-optimized after each ECM's movement, leading to better tracking and successful needle regrasps.
    }
    \label{fig:ECM-moves-needle-fixed-first-person}
    \vspace{-2.5mm}
\end{figure}

Table~\ref{tab:real-robot-success-rate} shows a comparison between the success rate of the baseline and \methodShortName{}, and Figs.~\ref{fig:cover_figure}, \ref{fig:ECM-fixed-needle-moves-first-person}, and \ref{fig:ECM-moves-needle-fixed-first-person} visualize their trajectories. 
For Type 1 (easy) trajectories, the regrasping arm starts outside the FOV and ends at a desired pose close to the center of the FOV at an ideal depth. 
The baseline is still able to achieve $50\%$ success in this (easy) setting since tracking is more likely to recover as the arm reaches its goal. 
We increase the difficulty in Type 1 (medium) case by moving the desired pose towards the edges of the FOV. 
Type 1 (hard) further adds to the challenge with dim lighting. 
In both the medium and hard scenarios, the baseline fails to complete a single regrasp, and \methodShortName{} significantly outperforms the baseline. 
For Type 2 trajectories (Fig.~\ref{fig:ECM-fixed-needle-moves-first-person}), the baseline sometimes succeeds since the goal of the needle arm is near the center of the FOV. 
For Type 3 trajectories, since the ECM only moves once at the beginning, the motion noise it injects can occasionally be recovered by the baseline. 
However, for Type 4 trajectories (Figs.~\ref{fig:cover_figure} and \ref{fig:ECM-moves-needle-fixed-first-person}), since the ECM moves four times throughout the trajectory, the significant motion noise causes the baseline to fail completely. 
\methodShortName{}, on the other hand, consistently achieves higher success under different environmental conditions.

Overall, \methodShortName{} achieves an $82\%$ success rate out of 60 regrasps, while the baseline only achieves $18\%$. 
By optimizing the baseline trajectory to minimize uncertainty in state estimation, our method enables the surgical tool and needle tracking to recover from significant motion and observation noise and prevents failure. 

%% file: texfiles/conclusion.tex

This work presents \methodShortName{}, a trajectory optimization framework that minimizes the uncertainty of tool tracking for robust surgical task automation. 
By considering motion and observation uncertainties in surgical scenes, \methodShortName{} outputs a tool trajectory such that the distinguishable features on the tool are better visible to a camera. 
Through 60 real-world suture needle regrasps on the dVRK, we demonstrate that \methodShortName{} significantly outperforms a baseline and consistently achieves a higher success rate under dark lighting, poor visibility, and a moving camera. 


Although not demonstrated in this work, \methodShortName{} can be generalized to consider other factors that help with surgical task automation. 
This includes directly optimizing the camera's movement and enabling collision avoidance by integrating collision constraints through differentiable collision checkers such as~\cite{zhi2022diffco}. 
While our motion and observation models are motivated by previous work, \cite{ali2020supervised,eslamian2016towards,ji2018learning,eslamian2020development,moccia2023autonomous,ou2023robot,lin2023autonomous,reiter2012feature,ye2016real,hao2018vision,li2020super,richter2021robotic,lu2021super,dambrosia2024robust}, we acknowledge that these models may be imperfect. 
Future work can include better-accuracy models optimized through approaches such as~\cite{greenberg2023optimization}.

%% file: main.bbl
\begin{thebibliography}{10}
\providecommand{\url}[1]{#1}
\csname url@samestyle\endcsname
\providecommand{\newblock}{\relax}
\providecommand{\bibinfo}[2]{#2}
\providecommand{\BIBentrySTDinterwordspacing}{\spaceskip=0pt\relax}
\providecommand{\BIBentryALTinterwordstretchfactor}{4}
\providecommand{\BIBentryALTinterwordspacing}{\spaceskip=\fontdimen2\font plus
\BIBentryALTinterwordstretchfactor\fontdimen3\font minus \fontdimen4\font\relax}
\providecommand{\BIBforeignlanguage}[2]{{%
\expandafter\ifx\csname l@#1\endcsname\relax
\typeout{** WARNING: IEEEtran.bst: No hyphenation pattern has been}%
\typeout{** loaded for the language `#1'. Using the pattern for}%
\typeout{** the default language instead.}%
\else
\language=\csname l@#1\endcsname
\fi
#2}}
\providecommand{\BIBdecl}{\relax}
\BIBdecl

\bibitem{hubens2003performance}
G.~Hubens, H.~Coveliers, L.~Balliu, M.~Ruppert, and W.~Vaneerdeweg, ``A performance study comparing manual and robotically assisted laparoscopic surgery using the da vinci system,'' \emph{Surgical Endoscopy and other interventional techniques}, vol.~17, no.~10, pp. 1595--1599, 2003.

\bibitem{khubchandani2018geographic}
J.~A. Khubchandani, A.~M. Ingraham, V.~T. Daniel, D.~Ayturk, C.~I. Kiefe, and H.~P. Santry, ``Geographic diffusion and implementation of acute care surgery: an uneven solution to the national emergency general surgery crisis,'' \emph{JAMA surgery}, vol. 153, no.~2, pp. 150--159, 2018.

\bibitem{kazanzides2014open}
P.~Kazanzides, Z.~Chen, A.~Deguet, G.~S. Fischer, R.~H. Taylor, and S.~P. DiMaio, ``An open-source research kit for the da vinci{\textregistered} surgical system,'' in \emph{2014 IEEE international conference on robotics and automation (ICRA)}.\hskip 1em plus 0.5em minus 0.4em\relax IEEE, 2014, pp. 6434--6439.

\bibitem{iyer2013single}
S.~Iyer, T.~Looi, and J.~Drake, ``A single arm, single camera system for automated suturing,'' in \emph{2013 IEEE International Conference on Robotics and Automation}.\hskip 1em plus 0.5em minus 0.4em\relax IEEE, 2013, pp. 239--244.

\bibitem{sen2016automating}
S.~Sen, A.~Garg, D.~V. Gealy, S.~McKinley, Y.~Jen, and K.~Goldberg, ``Automating multi-throw multilateral surgical suturing with a mechanical needle guide and sequential convex optimization,'' in \emph{2016 IEEE international conference on robotics and automation (ICRA)}.\hskip 1em plus 0.5em minus 0.4em\relax IEEE, 2016, pp. 4178--4185.

\bibitem{pedram2020autonomous}
S.~A. Pedram, C.~Shin, P.~W. Ferguson, J.~Ma, E.~P. Dutson, and J.~Rosen, ``Autonomous suturing framework and quantification using a cable-driven surgical robot,'' \emph{IEEE Transactions on Robotics}, vol.~37, no.~2, pp. 404--417, 2020.

\bibitem{schwaner2021autonomous}
K.~L. Schwaner, I.~Iturrate, J.~K. Andersen, P.~T. Jensen, and T.~R. Savarimuthu, ``Autonomous bi-manual surgical suturing based on skills learned from demonstration,'' in \emph{2021 IEEE/RSJ International Conference on Intelligent Robots and Systems (IROS)}.\hskip 1em plus 0.5em minus 0.4em\relax IEEE, 2021, pp. 4017--4024.

\bibitem{richter2021autonomous}
F.~Richter, S.~Shen, F.~Liu, J.~Huang, E.~K. Funk, R.~K. Orosco, and M.~C. Yip, ``Autonomous robotic suction to clear the surgical field for hemostasis using image-based blood flow detection,'' \emph{IEEE Robotics and Automation Letters}, vol.~6, no.~2, pp. 1383--1390, 2021.

\bibitem{huang2021model}
J.~Huang, F.~Liu, F.~Richter, and M.~C. Yip, ``Model-predictive control of blood suction for surgical hemostasis using differentiable fluid simulations,'' in \emph{2021 IEEE International Conference on Robotics and Automation (ICRA)}.\hskip 1em plus 0.5em minus 0.4em\relax IEEE, 2021, pp. 12\,380--12\,386.

\bibitem{oh2023framework}
K.-H. Oh, L.~Borgioli, M.~Zefran, L.~Chen, and P.~C. Giulianotti, ``A framework for automated dissection along tissue boundary,'' \emph{arXiv preprint arXiv:2310.09669}, 2023.

\bibitem{hao2018vision}
R.~Hao, O.~{\"O}zg{\"u}ner, and M.~C. {\c{C}}avu{\c{s}}o{\u{g}}lu, ``Vision-based surgical tool pose estimation for the da vinci{\textregistered} robotic surgical system,'' in \emph{2018 IEEE/RSJ international conference on intelligent robots and systems (IROS)}.\hskip 1em plus 0.5em minus 0.4em\relax IEEE, 2018, pp. 1298--1305.

\bibitem{lu2021super}
J.~Lu, A.~Jayakumari, F.~Richter, Y.~Li, and M.~C. Yip, ``Super deep: A surgical perception framework for robotic tissue manipulation using deep learning for feature extraction,'' in \emph{2021 IEEE International Conference on Robotics and Automation (ICRA)}.\hskip 1em plus 0.5em minus 0.4em\relax IEEE, 2021, pp. 4783--4789.

\bibitem{richter2021robotic}
F.~Richter, J.~Lu, R.~K. Orosco, and M.~C. Yip, ``Robotic tool tracking under partially visible kinematic chain: A unified approach,'' \emph{IEEE Transactions on Robotics}, vol.~38, no.~3, pp. 1653--1670, 2021.

\bibitem{dambrosia2024robust}
C.~D'Ambrosia, F.~Richter, Z.-Y. Chiu, N.~Shinde, F.~Liu, H.~I. Christensen, and M.~C. Yip, ``Robust surgical tool tracking with pixel-based probabilities for projected geometric primitives,'' \emph{arXiv preprint arXiv:2403.04971}, 2024.

\bibitem{d2018automated}
C.~D'Ettorre, G.~Dwyer, X.~Du, F.~Chadebecq, F.~Vasconcelos, E.~De~Momi, and D.~Stoyanov, ``Automated pick-up of suturing needles for robotic surgical assistance,'' in \emph{2018 IEEE International Conference on Robotics and Automation (ICRA)}.\hskip 1em plus 0.5em minus 0.4em\relax IEEE, 2018, pp. 1370--1377.

\bibitem{chiu2021bimanual}
Z.-Y. Chiu, F.~Richter, E.~K. Funk, R.~K. Orosco, and M.~C. Yip, ``Bimanual regrasping for suture needles using reinforcement learning for rapid motion planning,'' in \emph{2021 IEEE International Conference on Robotics and Automation (ICRA)}.\hskip 1em plus 0.5em minus 0.4em\relax IEEE, 2021, pp. 7737--7743.

\bibitem{wilcox2022learning}
A.~Wilcox, J.~Kerr, B.~Thananjeyan, J.~Ichnowski, M.~Hwang, S.~Paradis, D.~Fer, and K.~Goldberg, ``Learning to localize, grasp, and hand over unmodified surgical needles,'' in \emph{2022 International Conference on Robotics and Automation (ICRA)}.\hskip 1em plus 0.5em minus 0.4em\relax IEEE, 2022, pp. 9637--9643.

\bibitem{miyasaka2020modeling}
M.~Miyasaka, M.~Haghighipanah, Y.~Li, J.~Matheson, A.~Lewis, and B.~Hannaford, ``Modeling cable-driven robot with hysteresis and cable--pulley network friction,'' \emph{IEEE/ASME Transactions on Mechatronics}, vol.~25, no.~2, pp. 1095--1104, 2020.

\bibitem{hwang2020efficiently}
M.~Hwang, B.~Thananjeyan, S.~Paradis, D.~Seita, J.~Ichnowski, D.~Fer, T.~Low, and K.~Goldberg, ``Efficiently calibrating cable-driven surgical robots with rgbd fiducial sensing and recurrent neural networks,'' \emph{IEEE Robotics and Automation Letters}, vol.~5, no.~4, pp. 5937--5944, 2020.

\bibitem{eslamian2016towards}
S.~Eslamian, L.~A. Reisner, B.~W. King, and A.~K. Pandya, ``Towards the implementation of an autonomous camera algorithm on the da vinci platform,'' in \emph{Medicine Meets Virtual Reality 22}.\hskip 1em plus 0.5em minus 0.4em\relax IOS Press, 2016, pp. 118--123.

\bibitem{ji2018learning}
J.~J. Ji, S.~Krishnan, V.~Patel, D.~Fer, and K.~Goldberg, ``Learning 2d surgical camera motion from demonstrations,'' in \emph{2018 IEEE 14th International Conference on Automation Science and Engineering (CASE)}.\hskip 1em plus 0.5em minus 0.4em\relax IEEE, 2018, pp. 35--42.

\bibitem{moccia2023autonomous}
R.~Moccia and F.~Ficuciello, ``Autonomous endoscope control algorithm with visibility and joint limits avoidance constraints for da vinci research kit robot,'' in \emph{2023 IEEE International Conference on Robotics and Automation (ICRA)}.\hskip 1em plus 0.5em minus 0.4em\relax IEEE, 2023, pp. 776--781.

\bibitem{zhong2019dual}
F.~Zhong, Y.~Wang, Z.~Wang, and Y.-H. Liu, ``Dual-arm robotic needle insertion with active tissue deformation for autonomous suturing,'' \emph{IEEE Robotics and Automation Letters}, vol.~4, no.~3, pp. 2669--2676, 2019.

\bibitem{ozguner2021visually}
O.~{\"O}zg{\"u}ner, T.~Shkurti, S.~Lu, W.~Newman, and M.~C. {\c{C}}avu{\c{s}}o{\u{g}}lu, ``Visually guided needle driving and pull for autonomous suturing,'' in \emph{2021 IEEE 17th International Conference on Automation Science and Engineering (CASE)}.\hskip 1em plus 0.5em minus 0.4em\relax IEEE, 2021, pp. 242--248.

\bibitem{fozilov2023towards}
K.~Fozilov, J.~Colan, K.~Sekiyama, and Y.~Hasegawa, ``Towards autonomous robotic minimally invasive surgery: A hybrid framework combining task-motion planning and dynamic behavior trees,'' \emph{IEEE Access}, 2023.

\bibitem{dharmarajan2023automating}
K.~Dharmarajan, W.~Panitch, M.~Jiang, K.~Srinivas, B.~Shi, Y.~Avigal, H.~Huang, T.~Low, D.~Fer, and K.~Goldberg, ``Automating vascular shunt insertion with the dvrk surgical robot,'' in \emph{2023 IEEE International Conference on Robotics and Automation (ICRA)}.\hskip 1em plus 0.5em minus 0.4em\relax IEEE, 2023, pp. 6781--6788.

\bibitem{LQG-MP}
J.~Van Den~Berg, P.~Abbeel, and K.~Goldberg, ``Lqg-mp: Optimized path planning for robots with motion uncertainty and imperfect state information,'' \emph{International Journal of Robotics Research}, vol.~30, no.~7, pp. 895--913, 2011.

\bibitem{van2012motion}
J.~Van Den~Berg, S.~Patil, and R.~Alterovitz, ``Motion planning under uncertainty using iterative local optimization in belief space,'' \emph{The International Journal of Robotics Research}, vol.~31, no.~11, pp. 1263--1278, 2012.

\bibitem{sunberg2018online}
Z.~Sunberg and M.~Kochenderfer, ``Online algorithms for pomdps with continuous state, action, and observation spaces,'' in \emph{Proceedings of the International Conference on Automated Planning and Scheduling}, vol.~28, 2018, pp. 259--263.

\bibitem{huang2005multi}
S.~Huang, N.~M. Kwok, G.~Dissanayake, Q.~P. Ha, and G.~Fang, ``Multi-step look-ahead trajectory planning in slam: Possibility and necessity,'' in \emph{Proceedings of the 2005 IEEE international conference on robotics and automation}.\hskip 1em plus 0.5em minus 0.4em\relax IEEE, 2005, pp. 1091--1096.

\bibitem{clemens2016evidential}
J.~Clemens, T.~Reineking, and T.~Kluth, ``An evidential approach to slam, path planning, and active exploration,'' \emph{International Journal of Approximate Reasoning}, vol.~73, pp. 1--26, 2016.

\bibitem{ali2020supervised}
S.~Ali, Y.~Jonmohamadi, Y.~Takeda, J.~Roberts, R.~Crawford, and A.~K. Pandey, ``Supervised scene illumination control in stereo arthroscopes for robot assisted minimally invasive surgery,'' \emph{IEEE Sensors Journal}, vol.~21, no.~10, pp. 11\,577--11\,587, 2020.

\bibitem{eslamian2020development}
S.~Eslamian, L.~A. Reisner, and A.~K. Pandya, ``Development and evaluation of an autonomous camera control algorithm on the da vinci surgical system,'' \emph{The International Journal of Medical Robotics and Computer Assisted Surgery}, vol.~16, no.~2, p. e2036, 2020.

\bibitem{ou2023robot}
Y.~Ou, S.~Zargarzadeh, and M.~Tavakoli, ``Robot learning incorporating human interventions in the real world for autonomous surgical endoscopic camera control,'' \emph{Journal of Medical Robotics Research}, vol.~8, no. 03n04, p. 2340004, 2023.

\bibitem{lin2023autonomous}
H.-C. Lin, M.~M. Marinho, and K.~Harada, ``Autonomous field-of-view adjustment using adaptive kinematic constrained control with robot-held microscopic camera feedback,'' \emph{arXiv preprint arXiv:2309.10287}, 2023.

\bibitem{reiter2012feature}
A.~Reiter, P.~K. Allen, and T.~Zhao, ``Feature classification for tracking articulated surgical tools,'' in \emph{Medical Image Computing and Computer-Assisted Intervention--MICCAI 2012: 15th International Conference, Nice, France, October 1-5, 2012, Proceedings, Part II 15}.\hskip 1em plus 0.5em minus 0.4em\relax Springer, 2012, pp. 592--600.

\bibitem{ye2016real}
M.~Ye, L.~Zhang, S.~Giannarou, and G.-Z. Yang, ``Real-time 3d tracking of articulated tools for robotic surgery,'' in \emph{Medical Image Computing and Computer-Assisted Intervention--MICCAI 2016: 19th International Conference, Athens, Greece, October 17-21, 2016, Proceedings, Part I 19}.\hskip 1em plus 0.5em minus 0.4em\relax Springer, 2016, pp. 386--394.

\bibitem{li2020super}
Y.~Li, F.~Richter, J.~Lu, E.~K. Funk, R.~K. Orosco, J.~Zhu, and M.~C. Yip, ``Super: A surgical perception framework for endoscopic tissue manipulation with surgical robotics,'' \emph{IEEE Robotics and Automation Letters}, vol.~5, no.~2, pp. 2294--2301, 2020.

\bibitem{platt2010belief}
R.~Platt~Jr, R.~Tedrake, L.~P. Kaelbling, and T.~Lozano-Perez, ``Belief space planning assuming maximum likelihood observations.'' in \emph{Robotics: Science and Systems}, vol.~2, 2010.

\bibitem{chiu2022markerless}
Z.-Y. Chiu, A.~Z. Liao, F.~Richter, B.~Johnson, and M.~C. Yip, ``Markerless suture needle 6d pose tracking with robust uncertainty estimation for autonomous minimally invasive robotic surgery,'' in \emph{2022 IEEE/RSJ International Conference on Intelligent Robots and Systems (IROS)}.\hskip 1em plus 0.5em minus 0.4em\relax IEEE, 2022, pp. 5286--5292.

\bibitem{chiu2023real}
Z.-Y. Chiu, F.~Richter, and M.~C. Yip, ``Real-time constrained 6d object-pose tracking of an in-hand suture needle for minimally invasive robotic surgery,'' in \emph{2023 IEEE International Conference on Robotics and Automation (ICRA)}.\hskip 1em plus 0.5em minus 0.4em\relax IEEE, 2023, pp. 4761--4767.

\bibitem{paszke2019pytorch}
A.~Paszke, S.~Gross, F.~Massa, A.~Lerer, J.~Bradbury, G.~Chanan, T.~Killeen, Z.~Lin, N.~Gimelshein, L.~Antiga \emph{et~al.}, ``Pytorch: An imperative style, high-performance deep learning library,'' \emph{Advances in neural information processing systems}, vol.~32, 2019.

\bibitem{wang2023pypose}
C.~Wang, D.~Gao, K.~Xu, J.~Geng, Y.~Hu, Y.~Qiu, B.~Li, F.~Yang, B.~Moon, A.~Pandey \emph{et~al.}, ``Pypose: A library for robot learning with physics-based optimization,'' in \emph{Proceedings of the IEEE/CVF Conference on Computer Vision and Pattern Recognition}, 2023, pp. 22\,024--22\,034.

\bibitem{liu1989limited}
D.~C. Liu and J.~Nocedal, ``On the limited memory bfgs method for large scale optimization,'' \emph{Mathematical programming}, vol.~45, no.~1, pp. 503--528, 1989.

\bibitem{HJReachability}
S.~Bansal, M.~Chen, S.~Herbert, and C.~J. Tomlin, ``Hamilton-jacobi reachability: A brief overview and recent advances,'' 2017.

\bibitem{GP_reachability}
A.~K. Akametalu, J.~F. Fisac, J.~H. Gillula, S.~Kaynama, M.~N. Zeilinger, and C.~J. Tomlin, ``Reachability-based safe learning with gaussian processes,'' in \emph{53rd IEEE Conference on Decision and Control}, 2014, pp. 1424--1431.

\bibitem{cheng2023putting}
H.~K. Cheng, S.~W. Oh, B.~Price, J.-Y. Lee, and A.~Schwing, ``Putting the object back into video object segmentation,'' \emph{arXiv preprint arXiv:2310.12982}, 2023.

\bibitem{kirillov2023segment}
A.~Kirillov, E.~Mintun, N.~Ravi, H.~Mao, C.~Rolland, L.~Gustafson, T.~Xiao, S.~Whitehead, A.~C. Berg, W.-Y. Lo \emph{et~al.}, ``Segment anything,'' \emph{arXiv preprint arXiv:2304.02643}, 2023.

\bibitem{omote1999self}
K.~Omote, H.~Feussner, A.~Ungeheuer, K.~Arbter, G.-Q. Wei, J.~R. Siewert, and G.~Hirzinger, ``Self-guided robotic camera control for laparoscopic surgery compared with human camera control,'' \emph{The American journal of surgery}, vol. 177, no.~4, pp. 321--324, 1999.

\bibitem{zhi2022diffco}
Y.~Zhi, N.~Das, and M.~Yip, ``Diffco: Autodifferentiable proxy collision detection with multiclass labels for safety-aware trajectory optimization,'' \emph{IEEE Transactions on Robotics}, vol.~38, no.~5, pp. 2668--2685, 2022.

\bibitem{greenberg2023optimization}
I.~Greenberg, N.~Yannay, and S.~Mannor, ``Optimization or architecture: How to hack kalman filtering,'' 2023.

\end{thebibliography}
